# Joint Structured Learning and Predictions under Logical Constraints in Conditional Random Fields

Jean-Luc Meunier

Xerox Research Centre Europe

May 30, 2017.

## Abstract

This paper is concerned with structured machine learning, in a supervised machine learning context. It discusses how to make joint structured learning on interdependent objects of different nature, as well as how to enforce logical constraints when predicting labels.

We explain how this need arose in a Document Understanding task. We then discuss a general extension to Conditional Random Fields (CRF) for this purpose and present the contributed open source implementation on top of the open source PyStruct library. We evaluate its performance on a publicly available dataset.

**Keywords:** supervised machine learning, structured prediction, conditional random fields.

## 1 Introduction

Structured prediction is about predicting a structured output rather than a scalar value, given some input. It finds applications in natural language processing, computer vision and many other fields. While there are several models and relevant methods to support structured machine learning, our focus here is on the popular graphical model named Conditional Random Fields (CRF).

CRF was first introduced by Lafferty et al. in 2001 [1] and later applied on many tasks in computer vision, natural language processing to cite a few. We review here only a few salient works, relevant to the topic of this paper.

In 2004, Quattoni et al. [13] aimed at categorizing an image based on its parts. Since the parts were not labelled, i.e. unobserved, they introduced hidden variables. While this incorporation of hidden variables in the CRF framework is the key contribution, we also view this work as one of the first at dealing with objects of different natures in CRF. Indeed, the so-called parts are patches while it is the image itself that needs to be categorized. Similarly, He et al. [14] introduced an additional layer of hidden variables to encode particular patterns within a subset of nodes.

In 2004 also, Sutton et al. [15] introduced the concept of Factorial CRF, which is a generalization of linear-chain CRFs that repeat structure and parameters over a sequence of state vectors. In this approach, the labels of a chain belong each to a certain temporality, with connection between labels that are consecutive within-chain and between consecutive chains. One advantage of this model is to jointly solve multiple labelling tasks, on the same sequence, instead of applying multiple separate models. Recently, for instance, Wang & Kan in [16] used Factorial CRF to jointly perform Chinese word recognition and segmentation. We will show in section 3.2.2 how our proposal generalizes Factorial CRFs.

CRF models came later in the Document Understanding community and was used to segment and label a page image in [17]. Recently, and closer to the task discussed in next section, CRF was used for labeling the objects on a page of a document [19], moving away from pixel-based or pixel-patch-labeling.

The work of Albert et al. in [18] is also of particular relevance to us. They propose a two-layer CRF model to simultaneously classify the so-called *land cover* and *land use*, the former relates to the nature of the terrain while the latter relates to its socio-economic function. Both layers consist of nodes and intra-layer edges. The nodes at different layers are connected by inter-layer edges. *Both layers differ with respect to the entities corresponding to the nodes and the classes to be distinguished, which is caused by the different nature of these classification tasks.*

We are addressing here this problem, i.e. having nodes of different nature, in a general principled manner.

The goal of this paper it to focus on an extension of CRF supporting joint prediction on items of different nature but possibly interdependent. We also propose to support logical constraints on items' labelling. We illustrate the approach by an open source extension to the open source CRF library called PyStruct [2,3,4].

In the next sections, we will illustrate similar needs arising in the Document Understanding field. We then discuss how to support nodes of different nature in CRF and how to support first order logic constraints at inference time in PyStruct. Finally, we extend the Snake example introduced in [6] to experimentally validate this work.

## 2 Need for Joint Structured Prediction and Logical Constraints

### 2.1 Nodes of Different Natures

During a Document Understanding task consisting in labeling textual blocks of OCRed or digital documents, a CRF-based structured learning approach was chosen. We modelled each document as a general CRF graph, consisting only of unary and pairwise potential functions. In this graph, a node reflects a textual block while an edge reflects a spatial relation between two blocks, either on same page or on two contiguous pages [5]. The unary potential applies onto nodes, while the pairwise applies onto edges. Modeling the whole document as a graph allowed us to leverage the patterns that exist within a document, within and across its pages.

However, a secondary task consisted in labeling each page of a document. Since the page labels clearly follow a pattern over the sequence of pages of each document, we again used structured learning, a chain CRF model in this case.

Eventually, a structured classifier at block-level and a structured classifier at page-level were delivered for production use, using PyStruct.

But this two-part solution cannot capture and exploit the dependency between the two tasks. Our intuition is that the category of a page is informative for labelling its text blocks, and vice versa. We therefore believe a better solution could be constructed by addressing both tasks jointly instead of independently of each other. This would involve representing the text blocks and the pages of a document as nodes, of different types, in the same graph. Each type of node would have its own label space and edges would be allowed between pairs of nodes of same type and of different types. This was not possible with PyStruct, which on the contrary assumes that all nodes are homogeneous, that is they all have the same meaning. In other words, all nodes must have the same number of classes and these classes mean the same things. So, all nodes share the same weights (per class) and all edges share the same weights (per pair of class), as discussed later.

Beyond this case, CRF seems promising for labelling page objects simultaneously while exploiting their interdependences. However, objects on a page are of different nature, like text, image, graphical line, to mention a few conventional page objects. Each nature of object calls for a specific vector representation and a specific set of classes (labels).

Therefore, we propose CRF models where nodes can be of one of several possible natures, or *type*. This is discussed in next section in more details.

### 2.2 First order Logic Constraints

During the same work, several known facts were not explicitly expressed in our block model. In this case, it was given as fact that each page could have zero or one so-called 'number' and zero or one so-called 'title'. While the CRF model could learn that from the training set, having the capability to express this *a priori* knowledge in the model is beneficial. Given that one inference library used by PyStruct [3] has the capability of making inferences while taking into account certain first order logic constraints, we extended PyStruct to benefit from this mechanism, as discussed in next section.

## 3 Heterogeneous CRF Models

We first recall the definition of a CRF model before defining the proposed model.

### 3.1 Conditional Random Fields

The original CRF model [1] assumes that nodes are homogeneous, that is, they share the same label set. The prediction of a CRF model [1] is multivariate, i.e. it is a vector of length $n$ of discrete labels in space $\{1, ..., l\}$.

$$Y = \{1, ..., l\}^n$$

Let $G = (V, E)$ where V denotes the set of vertices, and E $\subset$ V $\times$ V the set of edges. The graph potential function $g(x, y)$ between an input $x$ and its structured label vector $y$ takes the form below.

$$g(x,y) = \sum_{v \in V} \psi_v(x, y_v) + \sum_{(v,w) \in E} \psi_{v,w}(x, y_v, y_w) \quad (1)$$

Here $\psi_v(x, y_v)$ is the unary potential function, while $\psi_{v,w}(x, y_v, y_w)$ is the pairwise potential functions.

Under the usual assumption that the potential functions are linear, like in PyStruct, $g(x,y)$ is:

$$g(x,y) = \sum_{v \in V} \theta_{y_v}^T \cdot \phi_V(v) + \sum_{(v,w) \in E} \vartheta_{y_v, y_w}^T \cdot \phi_E(v,w) \quad (2)$$

Here, $\theta_{y_v}$ is the model's unary weight vector given a node label $y_v$, $\phi_V(v)$ is a vector representation of the vertex $v$, $\vartheta_{y_v,y_w}$ is a pairwise weight vector given a pair of labels, and $\phi_E(v,w)$ is a vector representation of an edge $(v,w)$.

### 3.2 Heterogenous Nodes

We generalize the supported graphs to a form which is close to what is called multi-type graphs in [7]:

Let $G = (\cup_{t=1}^k V_t, \cup_{t=1}^k \cup_{t'=1}^k E_{t,t'})$ be a *k-type* graph where $V_t$ denotes a set of vertices of type $t$, and $E_{t,t'}$ denotes the set of edges connecting a vertex of type $t$ to a vertex of type $t'$. Note that $E_{t,t'}$ can be empty if none of the $t$-type vertices is connected to $t'$-type vertices.

We then consider $k$ label spaces of the form $\{1, \dots, l_t\}$ where $l_t$ is the number of labels of nodes of type $t \in \{1, \dots, k\}$.

We can generalize the PyStruct potential function to:

$$g(x,y) = \sum_{t=1}^k \sum_{v \in V_t} \theta_{y_v}^{t\ T} \cdot \phi_{V_t}(v) + \sum_{t=1}^k \sum_{t'=1}^k \sum_{(v,w) \in E_{t,t'}} \vartheta_{y_v, y_w}^{t,t'\ T} \cdot \phi_{E_{t,t'}}(v,w) \quad (3)$$

Here, each type $t$ defines its own unary weight vector $\theta_{y_v}^t$ given a label $y_v \in \{1, \dots, l_t\}$ and defines its own vector representation $\phi_{V_t}(v)$ of a vertex $v \in V_t$. Similarly, each pair of type $t, t'$ defines its own pairwise weight vector $\vartheta_{y_v, y_w}^{t,t'}$ given a pair of labels $y_v, y_w \in \{1, \dots, l_t\} \times \{1, \dots, l_{t'}\}$ and its own vector representation $\phi_{E_{t,t'}}(v,w)$ of the edge $(v,w) \in E_{t,t'}$.

Operationally, the unary and pairwise weights of single-type CRF in Equation (2) can be factorized into Equation below.

$$g(x,y) = \sum_{i=1}^l \theta_i^T \cdot \left[ \sum_{v_i \in \{v \in V | y_v = i\}} \phi_V(v_i) \right] + \sum_{i=1}^l \sum_{j=1}^l \vartheta_{i,j}^T \cdot \left[ \sum_{\substack{v_i \in \{v \in V | y_v = i\} \\ v_j \in \{v \in V | y_v = j\}}} \phi_E(v_i, v_j) \right] \quad (4)$$

By concatenating the $\theta_i$ and the $\vartheta_{i,j}$, on one side, into a single vector $\theta$ and, on the other side, concatenating with proper shift the feature vectors aggregated by label, or pair of labels, into a single vector $\phi(x,y)$, one gets an equation like:

$$g(x,y) = \theta \cdot \phi(x,y) \quad (5)$$

This is the form internally used by PyStruct, which then uses a learner method to learn the weights $\theta$ and an inference method to predict the label vector $y$ that maximizes the potential given the input $x$.

The generalized form for multi-type CRF graphs is amenable to a similar form, shown below.

$$g(x,y) = \sum_{t=1}^k \sum_{i=1}^{l_t} \theta_i^{t\ T} \cdot \left[ \sum_{v_i \in \{v \in V_t | y_v = i\}} \phi_{V_t}(v_i) \right] + \sum_{t=1}^k \sum_{t'=1}^k \sum_{i=1}^{l_t} \sum_{j=1}^{l_{t'}} \vartheta_{i,j}^{t,t'\ T} \cdot \left[ \sum_{\substack{v_i \in \{v \in V_t | y_v = i\} \\ v_j \in \{v \in V_{t'} | y_v = j\}}} \phi_{E_{t,t'}}(v_i, v_j) \right] \quad (6)$$

By concatenating the $\theta_i^t$ and the $\vartheta_{i,j}^{t,t'}$, we get the multi-type model weight vector $\theta$ while the aggregated feature vectors can be concatenated with proper shift to obtain the $\phi(x,y)$. We again end with Equation (5) and fit with the general PyStruct design.

This is how we implemented the multi-type CRF model in PyStruct, but the approach is quite general and applies to CRF models based on linear potential functions.

This implementation however requires an adaptation of the inference method, since nodes do not have all the same set of labels. We chose to adapt the AD3 [10,12] inference library for that purpose. AD3 is one of the inference method supported by PyStruct, it provides an "approximate maximum a posteriori (MAP) inference on factor graphs, based on the alternating directions method of multipliers". The adaptation is immediate as a CRF vertex $v \in V_t$ is reflected as a factor graph variable associated to a potential vector of length $l_t$, computed as $\theta^t . \phi_{V_t}(v_i)$. A CRF edge $(v, w) \in E_{t,t'}$ is reflected as a pairwise factor associated to a potential matrix of shape $l_t \times l_{t'}$ computed as $\vartheta^{t,t'} . \phi_{E_{t,t'}}(v, w)$.

The contributed source code is available in GitHub [8,9].

### 3.2.1 Number of Model Parameters

In unstructured machine learning, dealing with multi-type input is usually dealt with by training one classifier per type of input and associated set of labels. Alternatively, a single, but a priori sub-optimal, model could be set up by simple concatenation of the sets of labels in a single set, and concatenation of the corresponding features vectors with appropriate shift per type.

But in structured machine learning, more precisely in a CRF-based approach, doing so would not be effective because it would lead to learn useless model parameters, for irrelevant states given a node, and for irrelevant pairwise interaction between pair of states given a pair of node. Due to the regularization, those parameters should be set to 0 but in any case, such an approach would lead to an unnecessary large model, with a number of parameters equal to:

$$k.\sum_{t=1}^{k} l_t + (\sum_{t=1}^{k} l_t).(\sum_{t=1}^{k} l_t) \qquad (7)$$

In addition, one is not guaranteed against inconsistent prediction, given the type of a node. In such case, no easy workaround exists, as choosing among relevant labels even for a small number of nodes is quickly intractable.

The proposed approach lead to a smaller number of model parameters equal to:

$$\sum_{t=1}^{k} l_t + \sum_{t=1}^{k}\sum_{t'=1}^{k} l_t.l_{t'} \qquad (8)$$

### 3.2.2 Relationship with Factorial CRF

We reproduce below a figure from [15] showing in a graphical representation how the multiple labels ($y$ and $w$) interact in a Factorial CRF.

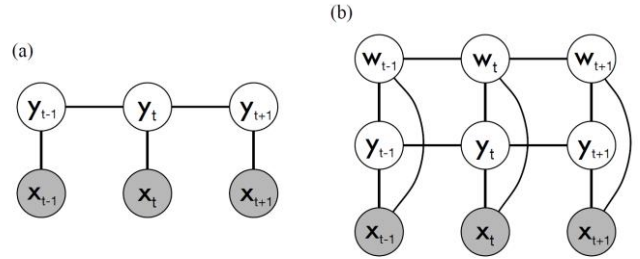

*Figure 1.* Graphical representation of (a) linear-chain CRF, and (b) factorial CRF. Although the hidden nodes can depend on observations at any time step, for clarity we have shown links only to observations at the same time step.

**Fig. 1.** Figure from [15]

The same representation is obtained in the present work by defining two types of nodes, one with labels $y$ and the other with labels $w$. Then each observation is reflected twice, once per node type, and by constructing a ladder-shape graph, one obtains the unary and pairwise interactions depicted in figure above. But contrary to Factorial CRF, other graph structures are possible as well, and in this sense, the notion of node type is more general.

Of course, as a practical remark, the representation of each observation does not need to be duplicated in memory, despite leading to two nodes in the graph.

### 3.3 First Order Logic Constraints

When using AD3 for inference, PyStruct converts the input CRF graph into a pairwise factor graph, as discussed in section 4.1.1 of Müller's thesis [4]. Noticing that the AD3 inference library can consider certain first order logic constraints for factor graphs made of binary variables and that, in addition, Martins et al. describe a procedure for reflecting CRF graphs as binary factor graphs, we considered supporting first order logic constraints expressed on the vertices' states of the CRF graph at inference time.

### 3.3.1 Logic Constraints in Single-Type PyStruct CRF

For supporting logic constraints in a single-type CRF, the CRF graph must be reflected as a binary pairwise factor graph. This is described in section 4.3 of [10] as follow:

- For each node $v \in V$ of a CRF graph, define binary variable $U_{v,i}$ for each possible label $i \in \{1, ..., l\}$; link those variables to a *XOR* factor. This imposes $\sum_{i=1}^{l} U_{v,i} = 1, \forall v \in V$.
- For each edge $(v,w) \in E$, define binary variables $U_{v,w,i,j}$ for each possible pair of labels $(i,j) \in \{1, ..., l\} \times \{1, ..., l\}$; link variables $\{U_{v,w,i,j}\}_{j=1}^{l}$ and $\neg U_{v,i}$ to a *XOR* factor for each $i \in \{1, ..., l\}$; and link variables $\{U_{v,w,i,j}\}_{i=1}^{l}$ and $\neg U_{w,j}$ to a *XOR* factor for each $j \in \{1, ..., l\}$. These impose constraints $U_{v,i} = \sum_{j=1}^{l} U_{v,w,i,j} \; \forall i$, and $U_{w,j} = \sum_{i=1}^{l} U_{v,w,i,j} \; \forall j$.

This procedure reflects each possible state of each node as a binary variable. It is straightforward to support first order logic constraints expressed over vertices' states at inference time, simply by adding those constraints. Obviously, adding such constraint(s) much be done with care so that the system is satisfiable.

The logic operators supported by AD3 are: XOR, XOR_OUT, AT_MOST_ONE, OR, OR_OUT, AND_OUT, IMPLY. Their semantic is detailed in section 5.2 of [12]. It is worth noting that XOR, OR, IMPLY are hard constraints, while XOR_OUT, OR_OUT, AND_OUT are soft-constraints. The former contributes $-\infty$ to the potential to while the latter rewards for fulfilling the constraint.

### 3.3.2 Logic Constraints in Multi-Type PyStruct CRF

A simple extension of the above procedure supports a $k$ types CRF graph. One has to build the corresponding pairwise binary factor graph as follow:

- For each type $t \in \{1, ..., k\}$, for each node $v \in V_t$ of a *k-type* CRF graph, define binary variable $U_{v,i}$ for each possible label $i \in \{1, ..., l_t\}$; link those variables to a *XOR* factor. This imposes $\sum_{i=1}^{l_t} U_{v,i} = 1, \forall v \in V_t, \forall t \in \{1, ..., k\}$.
- For each pair of type $(t, t')$, for each edge $(v,w) \in E_{t,t'}$, define binary variables $U_{v,w,i,j}$ for each possible pair of labels $(i,j) \in \{1, ..., l_t\} \times \{1, ..., l_{t'}\}$; link variables $\{U_{v,w,i,j}\}_{j=1}^{l_{t'}}$ and $\neg U_{v,i}$ to a *XOR* factor for each $i \in \{1, ..., l_t\}$; and link variables $\{U_{v,w,i,j}\}_{i=1}^{l_t}$ and $\neg U_{w,j}$ to a *XOR* factor for each $j \in \{1, ..., l_{t'}\}$. These impose constraints $U_{v,i} = \sum_{j=1}^{l_{t'}} U_{v,w,i,j} \; \forall i$, and $U_{w,j} = \sum_{i=1}^{l_t} U_{v,w,i,j} \; \forall j$.

We implemented the support for first order logic constraint for both single-type and multi-type CRF models in an inference method declared as 'ad3+' in PyStruct. The contributed source code is available in GitHub [8,9]. It consists in modifying the Python API of the AD3 library, which itself remains unchanged.

## 4 Experiment: Hidden Snakes

We experimented with our Document Understanding task the impact of the logic constraints and it showed some clear improvement. Also, we could train our CRF model over 584 documents, totalizing 5,173,073 nodes and 11,052,531 edges. With 2041 node features and 1034 edge features, and for 3 classes, this leads to a model containing 2041×3 + 1034×3×3 = 15429 parameters. The training took 5 days on a 2x Xeon E5-2680 @2.4Ghz computer. As PyStruct supports parallel computing, we allowed for up to 8 jobs in parallel, which shared some memory. In total, the memory footprint was ~400GB.

However, on this task, we did not experiment yet the multi-type CRF model and more importantly, we cannot provide the source code nor the data to let others reproduce the experiment.

Therefore, we report here on another task, which is artificial but reproducible but all.

We chose to reuse the *Snake* dataset introduced by Nowozin et al. in [6] and which was also used as one experiment by Müller in PyStruct [3, 4].

Here the task description made by Müller in [4], page 58-59:

> *The snakes dataset is a synthetic dataset where samples are labeled 2D grids. It was introduced by Nowozin et al. [2011] to demonstrate the importance of learning conditional pairwise potentials. The dataset consists of "snakes" of length ten traversing the 2D grid. Each grid cell that was visited is marked as the snake heading out towards the top, bottom, left, or right, while unvisited cells are marked as background.*
>
> *The goal is to predict a labeling of the snake from "head" to "tail", that is, assigning numbers from zero to nine to the cells that are occupied by the snake. Figure*

5.3 illustrates the principle. Local evidence for the target label is weak except around head and tail, making this a challenging task, requiring strong pairwise potentials. The dataset is noise-free in the sense that given the above description, a human could easily produce the desired labeling without making any mistake. The dataset is also interesting as the model proposed by Nowozin et al. [2011] produced notoriously hard-to-optimize energy functions.

Originally, the input is encoded into five RGB colors ("up", "down", "left", "right", "background"). To encode the input more suitably for our linear methods, we convert this representation to a one-hot encoding of the five states. We use a grid CRF model for this task. Unary potentials for each node are given by the input of the 8-neighborhood of the node—using a 4-neighborhood would most likely yield better results, but we do not want to encode too much task-knowledge into our model. Using the one-hot encoding of the input, this leads to $9 \cdot 5 = 45$ unary features. With 11 output classes, the unary potential has $11 \cdot 9 \cdot 5 = 495$ parameters. Features for the pairwise potentials are constructed by concatenating the features of the two neighboring nodes, taking the direction of the edge into account. The pairwise feature therefore has dimensionality $45 \cdot 4$, with the first 45 entries corresponding to the feature of the "top" node, the second 45 entries to the features "bottom" node, followed by the "left" and "right" nodes. As each edge is either horizontal or vertical, only two of these parts will be non-zero for any given edge. With $45 \cdot 4$ edge features, the pairwise potentials have $45 \cdot 4 \cdot 112 = 21780$ parameters.

To experiment the multi-type model as well as the logic constraints, we extended the task to obtain the *Hidden Snake* task. In this new task, we duplicate each image of the dataset and damage one of its snake cells by changing its color to another of the four colors among "*up*", "*down*", "*left*", "*right*". In most of the cases, the new image does no longer contain a snake of length ten but rather several shorter snake fragments. The label of all pixels of the modified image is set to 'background' since the image does not contain a snake of length ten, despite the 10 colored pixels. Let's note that in some surprising cases, despite the color change of one cell, we still have a snake of length ten. In this case, we discard the modified image. We therefore have slightly less images without, than with a snake. **Fig. 2** illustrates how a valid snake image is modified to obtain a 'no-snake' image.

We call this new dataset "Hidden Snake", since an image containing a snake looks like an image that does not contain one, as if the snake was hiding itself.

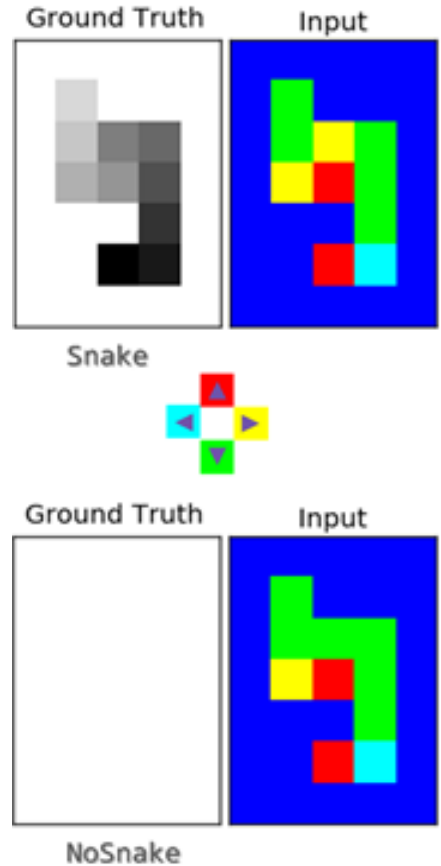

**Fig. 2.** The Hidden Snake task. Upper input image contains a snake, while the lower input image does not contain any snake and the groundtruth label of all its pixels is 'background', i.e. white colored. The legend of direction colors is shown in the middle.

Obviously, the task of labelling the snake cells from tail to head becomes more difficult, since only local evidence allow for detecting the snake discontinuity, which must lead to labelling all colored pixels as background.

In this experiment, **a secondary task** consists in labelling the image as Snake or NoSnake, depending if it contains a snake or not.

The original *Snake* dataset is available in GitHub. The code for modifying it as well. The size of the *Hidden Snake* dataset is reported in the **Table 1** below.

|          | *Snake* images | *NoSnake* images |
|----------|----------------|------------------|
| Training | 200            | 176              |
| Test     | 100            | 87               |

**Table 1.** Size of *Hidden Snake* dataset

### 4.1 Vector Representation and Hyper Parameters

In our experimentation, we kept the exact same vector representation of the pixels, used to solve the original Snake task. It is explained verbatim in the beginning of section 4.

The vector representation of the image itself is constituted of 7 values: the height and width in pixels of the non-background area in the image, and the histogram of number of pixels for each of the 5 colors. This representation is arbitrary and for sure not appropriate to solve the image classification task.

The hyper parameters of the Snake experiment by Müller are kept unchanged (at their default value, actually) for all experiments.

### 4.2 Experiments

We did three series of experiments. Firstly, to evaluate how harder is the *Hidden Snake* task compared to the original *Snake* task, using the standard PyStruct method. Secondly, we evaluate the performance of the extended PyStruct compared to the original PyStruct on the *Hidden Snake* task. Thirdly, we randomly generate additional images of snake to compare the same three methods.

In first two series, we also report the performance of a baseline oracle that predicts background for all pixels.

We did not compare PyStruct and the proposed extension on the *Snake* dataset, because by construction they are bound to perform the same, so this would be an engineering regression test, irrelevant here.

#### 4.2.1 How difficult is the *Hidden Snake* dataset?

The experiment consists in running the standard PyStruct method, which solved the *Snake* task, on the *Hidden Snake* task[1].

We report the global accuracy as done in previous work, but also the accuracy computed only on the snake cells, i.e. accuracy on labels 1 to 10 (label 0 is background).

This is shown in **Table 2**.

The Hidden Snake task is clearly more difficult and the performance of the CRF model degrades importantly, and comes close to the baseline 'all-background' oracle.

#### 4.2.2 Comparing to PyStruct on the Hidden Snake dataset

We trained and tested the methods below solely on the *Hidden Snake* dataset (the corresponding code is in GitHub[2]):

1. **PyStruct**: The PyStruct CRF model that was used to solve the original Snake task by Müller. It uses the grid CRF and the pixel representation vector. As explained verbatim, *features for the pairwise potentials are constructed by concatenating the features of the two neighboring nodes, taking the direction of the edge into account.*
   This is the baseline method.
2. **Logit**: A Logistic Regression model to classify an image as *Snake* or *NoSnake* solely based on the image vector representation.
3. **PyStruct+**: The extended PyStruct multi-type CRF model: we extend the grid CRF model used in method 1, by adding one additional node of a different type to reflect the image itself, with *Snake* or *NoSnake* as possible labels and the same vector representation as in method 2 for this type of node. We also create an edge from each pixel to the 'image' node, with a vector representation identical to the pixel representation vector.
4. **PyStruct+Logic**: here we do as in PyStruct+ (3 above) but at test time, we inject some logic constraints in the inference. We arbitrarily chose to express the fact that a snake has at most one cell of each possible cell label. So, for each test image, we create N constraints with N in [1,10], each constraint linking the state N of each pixel node of the CRF graph representing the image.

This is shown in **Table 3**.

The logit method performs badly, as expected because the features associated with an image are not sufficient at all to solve the task. This experiments shows the value of the multi-type model since we obtain a joint classification of both the pixels and the images, with better performance on each task compared to the corresponding specific classifiers. The logic constraints used at inference time are bringing additional performance gain at pixel-level, but equivalent loss at picture-level.

---

[1] In GitHub, files *examples/plot_snakes.py* and *examples/plot_hidden_snakes.py*

[2] In GitHub in file *examples/plot_hidden_short_snakes_typed.py*

### 4.2.3 Comparing to PyStruct on the Hidden Snake dataset with additional images

Here, instead of using the standard snake images, we generate them at random to evaluate the impact of the training set size to the methods. The test set is the same as before (the corresponding code is in GitHub[3]). Since we generate the training set, we ran the experiment ten times for each training set size.

This is shown in **Table 4**.

The average accuracy of the extended version of PyStruct is highly significantly above the one of PyStruct. However, the difference between PyStruct+ and PyStruct+Logic is not significant at 95% (Welch's t-tests), but only at 80%.

We believe this experiment validates the interest of the extension made on PyStruct.

|  |  | #pixels | Pixel accuracy | Pixel 1-10 accuracy |
|---|---|---|---|---|
| **Snake** | All-background oracle | 3750 | 0.733 | 0.000 |
|  | PyStruct | 3750 | **0.997** | **0.989** |
| **Hidden Snake** | All-background oracle | 7015 | 0.857 | 0.000 |
|  | PyStruct | 7015 | **0.917** | **0.803** |

**Table 2.** Performance of the single-type CRF model on the *Snake* and *Hidden Snake* datasets

| *Hidden Snake* dataset | #Pixels | Pixel accuracy | Pixel 1-10 accuracy | #Images | Image accuracy |
|---|---|---|---|---|---|
| **All-background Oracle** | 7015 | 0.86 | 0.00 | 187 | n/a |
| **PyStruct** | 7015 | 0.92 | 0.80 | 187 | n/a |
| **Logit** | 7015 | n/a | n/a | 187 | 0.44 |
| **PyStruct+** | 7015 | **0.94** | 0.87 | 187 | **0.84** |
| **PyStruct+Logic** | 7015 | **0.94** | **0.89** | 187 | 0.82 |

**Table 3.** Comparing PyStruct and its extension on the *Hidden Snake* dataset

| Nowozin test set, average accuracy on 10 runs ||||||| 
|---|---|---|---|---|---|---|
| **Generated train set size** | PyStruct || PyStruct + || PyStruct+Logic ||
|  | avg | stdev | avg | stdev | avg | stdev |
| 200 | 90.5% | 0.5% | 94.5% | 0.6% | **94.9%** | 0.7% |
| 400 | 93.3% | 0.3% | 96.6% | 0.5% | **97.0%** | 0.5% |
| 600 | 94.5% | 0.6% | 96.9% | 0.4% | **97.2%** | 0.4% |
| 800 | 95.3% | 0.4% | 97.3% | 0.4% | **97.5%** | 0.3% |

**Table 4.** Effect of larger training set

---

[3] In GitHub, file *examples.plot_hidden_short_snakes_typed_gen.py*

## 5  Conclusion

We presented an extension to CRF that support structured learning on objects of different nature, as well as prior knowledge in the form of logic constraints expressed on the predicted labels. This extension has been implemented on top of the PyStruct and to some extend the AD3 libraries, resulting in new open source libraries under either BSD or MIT license, both being very permissive.

We also proposed a modification of a dataset publicly available to evaluate the proposed extension.

We plan to conduct further Document Understanding experiments of this extension during the READ EU project on historical documents. We are also now looking forward from feedback from other practitioners.

## Acknowledgements

The author would like to thank his colleagues Jean-Marc Andréoli, Hervé Déjean, Matthias Gallé and Chunyang Xiao for their support and fruitful discussions, as well as Andreas Müller for his support on the PyStruct forum.

This work was funded by the EU project READ. The READ project has received funding from the European Union's Horizon 2020 research and innovation programme under grant agreement No 674943.## References